\documentclass[times,twoside]{zHenriquesLab-StyleBioRxiv}
\usepackage{blindtext}
\usepackage{textcomp}
\usepackage{floatrow}
\usepackage[label font=bf,labelformat=simple]{subfig}
\usepackage{subfig}
\usepackage{float}
\usepackage{overpic}
\usepackage{bbding}
\usepackage{array}
\usepackage{makecell}
\usepackage{bookmark}
\usepackage{subcaption}
\usepackage{caption}
\usepackage[colorinlistoftodos]{todonotes}
\usepackage{soul}
\usepackage[]{multibib}
\newcites{Method}{References}

\DeclareCaptionLabelSeparator{pipe}{ $\boldsymbol{|}$ }
\DeclareCaptionLabelFormat{supfiglabel}{\textbf{Extended Data Fig. #2}}
\DeclareCaptionType{supplementaryfigure}[Extended Data Fig.][List of Supplementary Figures]

\DeclareCaptionLabelFormat{suptablabel}{\textbf{Extended Data Table #2}}
\DeclareCaptionType{supplementarytable}[ Extended Data Table][List of Supplementary Tables]

\usepackage{graphicx}%
\usepackage{multirow}%
\usepackage{amsmath}
\usepackage{amssymb}
\usepackage{amsfonts}%
\usepackage{mathrsfs}%
\usepackage[title]{appendix}%
\usepackage{xcolor}%
\usepackage{textcomp}%
\usepackage{manyfoot}%
\usepackage{booktabs}%
\usepackage{algorithm}%
\usepackage{algorithmicx}%
\usepackage{algpseudocode}%
\usepackage{listings}%
\usepackage{hyperref}
\usepackage{wrapfig}
\usepackage{graphics}
\usepackage{epstopdf}
\usepackage{pifont}
\usepackage{url}
\leadauthor{Zhang}
\begin{document}
\title{AIBuildAI-2: A Knowledge-Enhanced Agent for Automatically Building AI Models}
\shorttitle{AIBuildAI-2}


\author[1, \textbf{*}]{Ruiyi Zhang}
\author[1, \textbf{*}]{Peijia Qin}
\author[1, \textbf{*}]{Qi Cao}
\author[1, \textbf{*}]{Li Zhang}
\author[1, 2, \Letter]{Pengtao~Xie}

\affil[1]{Department of Electrical and Computer Engineering, University of California San Diego, La Jolla, CA 92093, USA}
\affil[2]{Department of Medicine, University of California San Diego, La Jolla, CA 92093, USA}

\maketitle
\begin{abstract}
AI models underpin data-centric applications such as image processing, text processing, clinical decision support, and increasingly the acceleration of discovery across the natural sciences such as biology, physics, and chemistry. Yet developing them remains a heavily manual process in which practitioners must design architectures, build training pipelines, and iteratively refine solutions, and it can be challenging for natural scientists without specialized AI engineering expertise to build the high-performing models their research demands. To reduce this burden and broaden access to AI for scientific discovery, agents that automatically build AI models have been proposed. However, the performance of these agents is largely limited by the parametric knowledge of their underlying large language models, which is static, often outdated, and sparse on practical AI model engineering know-how. To address this limitation, we introduce AIBuildAI-2, a knowledge-enhanced agent with an external, evolving knowledge system for automatically building AI models. The knowledge system of AIBuildAI-2 is hierarchical, organizing curated AI development knowledge into high-level knowledge instructions over topical categories and low-level knowledge documents under each category, from which the agent dynamically loads only the context relevant to its current state and the AI task being solved, grounding each design and implementation decision in concrete, externally verifiable expertise. The system is initialized by collecting and cleaning AI-development-related documents from the web and organizing them into the corresponding categories, and continually evolves from the agent's own experience by distilling each completed run on an AI task into structured takeaways that are written back into the knowledge system. AIBuildAI-2 ranks first on MLE-Bench with a medal rate of 70.7\%, substantially exceeding all existing baseline methods. In a heart disease prediction competition, AIBuildAI-2 placed in the top 6.6\% among 4,370 teams composed of human experts. These results demonstrate that an effective knowledge system significantly improves the capabilities of AI model development agents to a level that meets that of expert human practitioners.
\end{abstract}

\begin{corrauthor}
\text{p1xie@ucsd.edu}
\end{corrauthor}

\begin{equal}
* Equal contribution
\end{equal}

\section*{Introduction}

AI models have become a foundational technology across modern data-centric applications, from general image and text processing~\cite{he2016deep,otter2020survey} to clinical decision support~\cite{singhal2023clinical,singhal2025medqa}. They are also increasingly used to accelerate discovery across the natural sciences such as biology, physics, and chemistry~\cite{jumper2021alphafold,wang2023scientific,stokes2020antibiotic,wong2023antibiotics}. In these settings, they have enabled data-driven analysis and decision-making at unprecedented scale~\cite{turing2007computing,jordan2015machine,ghahramani2015probabilistic,biamonte2017quantum}. Despite this progress, the development of competitive AI models for such tasks remains costly and reliant on specialized expertise: practitioners must iteratively formulate modeling strategies, implement training pipelines, optimize hyperparameters, and refine their approach in light of empirical evidence~\cite{bergstra2012random,sculley2015hidden,hutter2019automated,aldoseri2023re,yang2024limits}. In particular, it can be challenging for natural scientists without specialized AI engineering expertise to build the high-performing models their research demands. To reduce this burden and broaden access to AI for scientific discovery, a growing line of work has developed autonomous AI agents that, given only a task description and training data, can produce competitive AI models end-to-end. Recent agents based on large language models (LLMs)~\cite{toledo2025airesearchagentsmachine,Du2025AutoMLGenNF,jiang2025aide,yang2025rdagent,zhang2026aibuildai} have demonstrated promising performance on realistic AI development benchmarks such as MLE-Bench~\cite{chan2025mlebench}, and increasingly approach the capability of experienced AI engineers on a broad range of Kaggle-style~\cite{kaggle} tasks.
\newline
\newline
Despite this progress, the knowledge available to existing AI development agents during decision-making is fundamentally limited. The simplest agents rely solely on the parametric knowledge embedded in their underlying LLMs~\cite{jiang2025aide}, which is inherently static and often lacks the practical engineering know-how that distinguishes strong AI solutions from mediocre ones. Some agents augment this with on-the-fly web search to retrieve external information when the LLM is uncertain~\cite{toledo2025airesearchagentsmachine,zhang2026aibuildai}, but search results are noisy, inconsistent across queries, and rarely match the depth of curated technical references. Another line of work~\cite{Du2025AutoMLGenNF,yang2025rdagent} couples the agent with a fixed corpus of pre-collected documents, such as research papers and reports, providing more reliable context but only as a static snapshot that, given the rapid evolution of the AI field, cannot keep pace with newly emerging methods and best practices, and that cannot incorporate the agent's own accumulated experience across runs. Across all of these designs, the knowledge component is either too thin, too unreliable, or too static to consistently support the full breadth of decisions that arise in real AI model development, which limits how far autonomous agents can push the frontier of automated AI development.
\newline
\newline
To address this challenge, we introduce AIBuildAI-2, an autonomous AI development agent built around a comprehensive, hierarchical, and continually evolving knowledge system. The knowledge system has two levels: a top level of around 30 topical categories, defined by expert AI researchers, that span the AI model development lifecycle (e.g., specific tasks or modalities such as vision, NLP, and tabular data, and modeling strategies such as ensemble techniques), where each category carries a high-level knowledge instruction summarizing the strategies and best practices for that category; and a bottom level of around 1{,}000 low-level knowledge documents, to which each high-level knowledge instruction points for more specific knowledge. To remain efficient at this scale, AIBuildAI-2 employs dynamic context loading: every agent call always keeps the top-level index of categories in context, conditionally retrieves the high-level knowledge instructions whose categories are most relevant to the current AI task and decision, and fetches low-level knowledge documents only when finer-grained knowledge is needed for a specific design or implementation choice. The system is initialized by collecting and cleaning AI-development-related sources from the web, such as technical blogs, library documentation, and curated open-source code repositories, distilling them into the bottom-level knowledge documents, and synthesizing the high-level knowledge instruction for each category from these references. After each completed run, a knowledge-builder agent distills the experience into a structured takeaway that is appended to the bottom-level knowledge set under the most relevant category and used to incrementally update the corresponding high-level knowledge instruction, allowing the knowledge system to grow with the agent's experience.
\newline
\newline
We evaluate AIBuildAI-2 across three settings spanning standardized benchmarks and real-world competitions against human experts. On MLE-Bench~\cite{chan2025mlebench}, a benchmark of realistic Kaggle-style AI development tasks spanning visual, textual, time-series, and tabular modalities, AIBuildAI-2 ranks first on the leaderboard~\cite{mlebench_commit_2026} with a medal rate of 70.7\%, substantially outperforming all existing autonomous AI development systems including AIRA-dojo~\cite{toledo2025airesearchagentsmachine}, MLEvolve~\cite{Du2025AutoMLGenNF}, AIDE~\cite{jiang2025aide}, and R\&D-Agent~\cite{yang2025rdagent}. In a live heart disease prediction competition, AIBuildAI-2 placed in the top 6.6\% among 4,370 teams composed of human experts, demonstrating that an autonomous agent can match the performance of expert human practitioners on a clinically meaningful prediction task. In a drug discovery challenge, AIBuildAI-2 ranked in the top 38.8\% of human-expert participants, demonstrating broad applicability to scientifically demanding domains. Together, these results show that pairing a multi-agent system with a comprehensive, dynamically retrieved, and self-updating knowledge system enables AI agents to automate AI model development at a level that meets or exceeds expert human practitioners.

\section*{Results}
\subsection*{Overview of AIBuildAI-2}

\begin{figure*}[!htb]
\centering
\includegraphics[width=0.95\linewidth]{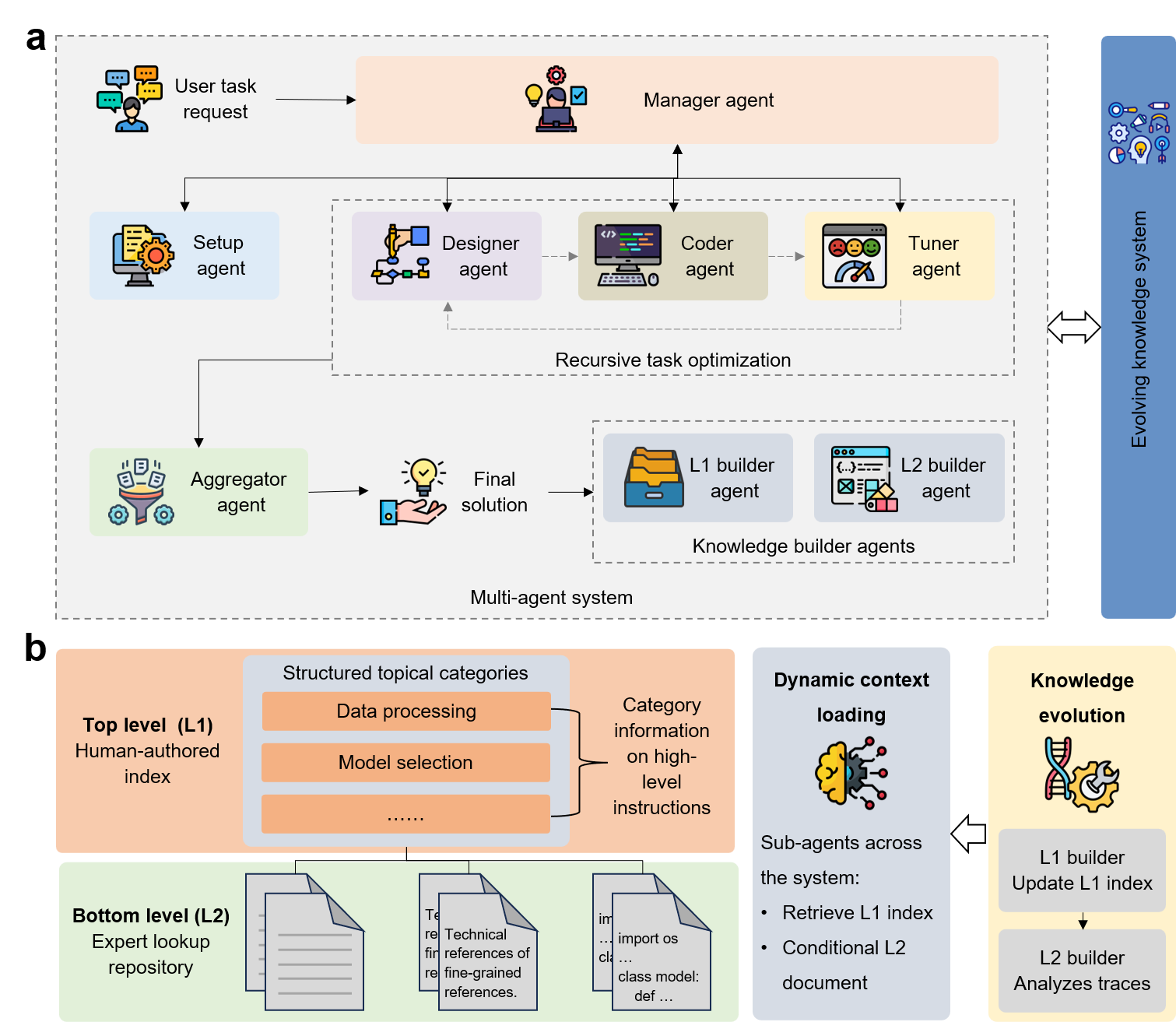}
\caption{\textbf{Overview of AIBuildAI-2.}
\textbf{a}, Overall workflow. AIBuildAI-2 takes as input a text description of an AI task and training data, and outputs trained model checkpoints and an inference script. It adopts a hierarchical sub-agent architecture: a setup agent prepares the software environment, a manager agent then oversees multiple parallel solution repositories and invokes a designer, coder, or tuner sub-agent on each, and an aggregator agent finally selects or ensembles the strongest repositories. All sub-agents are connected to a comprehensive, hierarchical, and continually evolving knowledge system, which they query for external expertise to ground their design, implementation, and tuning decisions. After each completed run, knowledge-builder agents distill the run into the knowledge system, so that it grows with the agent's experience.
\textbf{b}, Knowledge system. The knowledge system has two levels: a top level (L1), a human-authored index of around 30 topical categories of AI model development (e.g., data processing, model selection, and ensemble techniques), each carrying a high-level knowledge instruction; and a bottom level (L2), an expert lookup repository of around 1{,}000 low-level knowledge documents that the L1 instructions point to for more specific knowledge. Through dynamic context loading, sub-agents across the system retrieve the L1 index and conditionally load L2 documents. The knowledge system is evolved by two builder agents: an L2 builder that analyzes execution traces to produce new L2 documents, and an L1 builder that incrementally updates the L1 instructions.}
\label{fig:AIBuildAI_workflow}
\end{figure*}

AIBuildAI-2 is an AI agent that takes as input a text description of an AI task and training data, automatically designs, implements, and trains an AI model, and outputs the trained checkpoints and inference script for unseen test data. It adopts a hierarchical agent architecture similar to AIBuildAI~\cite{zhang2026aibuildai}, which mirrors real-world AI research workflows by coordinating a manager agent with three sub-agents: a designer for modeling and training strategy, a coder for translating designs into executable pipelines and iteratively debugging them, and a tuner for launching training runs and optimizing hyperparameters from logs. Each sub-agent performs multi-step reasoning and tool use across multiple LLM calls within a single invocation. The manager maintains multiple parallel solution repositories, each containing the model design, code, configurations, logs, and evaluation results, and decides which sub-agent to invoke on each repository through long-context reasoning over the accumulated execution history, refining promising candidates and discarding weak ones. A setup agent prepares the conda environment at initialization, and an aggregator agent selects or ensembles the strongest candidates at termination to produce the final submission-ready model (Fig.~\ref{fig:AIBuildAI_workflow}a).
\newline
\newline
The central methodological advance of AIBuildAI-2 is a comprehensive, hierarchical, and continually evolving knowledge system that grounds every design and implementation decision in concrete external expertise (Fig.~\ref{fig:AIBuildAI_workflow}b). The knowledge system is organized as two nested key--value lookups corresponding to two levels of knowledge: a top level (L1) of high-level knowledge instructions that summarize strategies and best practices for each topical category of AI model development, and a bottom level (L2) of low-level knowledge documents that provide more specific knowledge tailored to narrower scenarios within each category. The L1 index defines a human-authored taxonomy of around 30 topical categories, a large fraction of which target specific tasks or modalities such as vision, NLP, time-series, tabular, and molecular data, with the remainder covering general modeling strategies such as ensemble techniques. For each category, the corresponding L1 value contains the high-level knowledge instruction together with an index of pointers into the L2 lookup, which stores around 1{,}000 low-level knowledge documents, each providing more specific knowledge for a narrower scenario such as a feature extraction recipe for a particular type of data or tricks for improving a specific type of model.
\newline
\newline
To use this knowledge base efficiently within the limited context window of an LLM, AIBuildAI-2 employs a dynamic context-loading mechanism, in the spirit of agentic retrieval-augmented generation~\cite{lewis2020rag,singh2025agenticrag}. The L1 index is included in every agent call by default, providing a global map of available knowledge. Based on the current AI task description, the state of the active solution repository, and the role of the calling sub-agent, the manager and sub-agents conditionally retrieve only the L1 values whose categories are most relevant to the decision at hand, for example pulling in time-series modeling and tabular feature engineering categories for a forecasting task while ignoring image segmentation guidance. Low-level knowledge documents are retrieved only when an agent explicitly determines that more specific knowledge tailored to a narrower scenario (for example, a feature extraction recipe for a particular type of data or tricks for improving a specific type of model) is needed to make a particular design or implementation choice. This hierarchical loading strategy keeps the active context focused and concise even though the underlying knowledge base contains around 1{,}000 low-level knowledge documents.
\newline

\begin{figure*}[!htb]
\centering
\includegraphics[width=.75\linewidth]{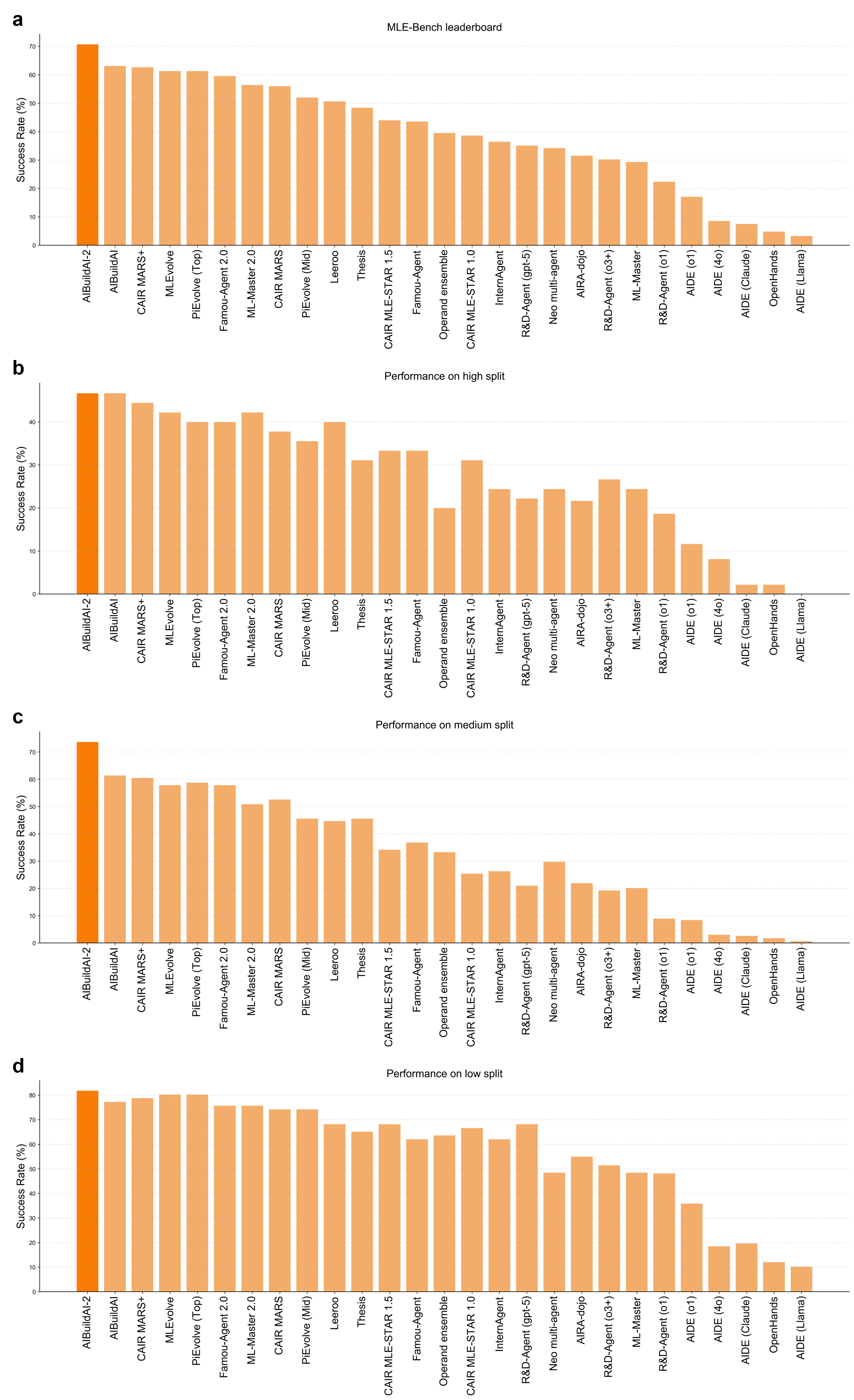}
\caption{\textbf{AIBuildAI-2 ranks first on the MLE-Bench leaderboard.}
\textbf{a}, Overall medal rate of AIBuildAI-2 compared with recent baseline methods (or method variants) on MLE-Bench. AIBuildAI-2 achieves the top overall position with a medal rate of 70.7\%, surpassing prior systems including MARS, Famou-Agent, ML-Master, Leeroo, InternAgent, R\&D-Agent, AIDE, AIRA-dojo, MLEvolve, and AIBuildAI.
\textbf{b--d}, Medal rates of AIBuildAI-2 and baseline methods on the high- (\textbf{b}), medium- (\textbf{c}), and low-complexity (\textbf{d}) splits of MLE-Bench. AIBuildAI-2 ranks first on all three splits, demonstrating consistent strong performance across tasks of varying difficulty.}
\label{fig:mle-bench}
\end{figure*}

The knowledge system is initialized by crawling, deduplicating, and cleaning AI-development-related sources from the web, including technical blogs, library and framework documentation, and curated open-source code repositories, and distilling them into the corresponding categories. To enable continual evolution thereafter, AIBuildAI-2 closes the loop between initial knowledge building and knowledge maintenance through two LLM-based builder agents. Once a run terminates, an L2 builder analyzes the full execution trace, including the design choices that were made, the experiments attempted, the failures encountered, and the strategies that produced the strongest solution, and generates a structured takeaway document that is appended to the L2 lookup as a new reference under the most relevant category. An L1 builder then uses this takeaway to incrementally update the corresponding L1 value whenever the run reveals new strategies, refinements, or failure modes that are not yet captured. The same two builders are also used to fold in newly published competition write-ups, blogs, library updates, and code repositories so that the knowledge base continues to track the state of the field. As more tasks are solved and more web content is incorporated, the knowledge system accumulates increasingly comprehensive and battle-tested guidance, allowing AIBuildAI-2 to leverage not only the collective expertise embedded in publicly available documents but also the experience of its own previous runs.

\subsection*{AIBuildAI-2 ranks first on the MLE-Bench leaderboard}

We evaluate AIBuildAI-2 on MLE-Bench, one of the most comprehensive benchmarks for assessing autonomous AI systems on realistic Kaggle-style tasks that require end-to-end AI model development rather than isolated algorithmic components~\cite{chan2025mlebench}. MLE-Bench contains 75 tasks curated from past Kaggle competitions, each providing raw datasets, evaluation metrics, and submission protocols faithful to real-world AI development, and requiring systems to design modeling strategies, implement training pipelines, and iteratively improve performance under limited computational budgets. Performance is measured using the medal system derived from human Kaggle competition rankings, where submissions that meet predefined percentile thresholds on the Kaggle leaderboard are awarded medals. As Kaggle competitions draw a global community of expert AI researchers and developers, earning a medal serves as a direct proxy for AI development capability on par with top human practitioners.
\newline
\newline
AIBuildAI-2 ranks first on the MLE-Bench leaderboard with a medal rate of 70.7\% (Fig.~\ref{fig:mle-bench}a), establishing a new state of the art for autonomous AI development. AIBuildAI-2 outperforms all recent baseline methods (or method variants) on the leaderboard, including MARS~\cite{chen2026mars}, Famou-Agent~\cite{li2025fmagent}, ML-Master~\cite{liu2025mlmaster}, Leeroo~\cite{nadaf2026kapso}, InternAgent~\cite{team2025internagent}, R\&D-Agent~\cite{yang2025rdagent}, AIDE~\cite{jiang2025aide}, AIRA-dojo~\cite{toledo2025airesearchagentsmachine}, MLEvolve~\cite{Du2025AutoMLGenNF}, and AIBuildAI~\cite{zhang2026aibuildai}. Together, these results demonstrate that AIBuildAI-2 matches the capability of top-tier human AI researchers and developers across a broad spectrum of real-world AI development tasks without any task-specific customization, highlighting the effectiveness of its knowledge-enhanced hierarchical sub-agent design for autonomous AI model development.
\newline
\newline
Fig.~\ref{fig:mle-bench}b--d further reports the performance of AIBuildAI-2 across the three complexity splits of MLE-Bench, comprising 15 high-complexity, 38 medium-complexity, and 22 low-complexity tasks, with complexity reflecting the difficulty of building a competitive solution. AIBuildAI-2 achieves medal rates of 46.67\%, 73.68\%, and 81.82\% on the high-, medium-, and low-complexity splits, respectively, ranking first on all three splits. The consistent top performance across difficulty levels highlights the generalizability and robustness of AIBuildAI-2 in autonomously building competitive AI models across a wide range of tasks.

\subsection*{AIBuildAI-2 reaches top-tier human-expert performance on a heart disease prediction competition}

\begin{figure*}[!htb]
\centering
\includegraphics[width=\linewidth]{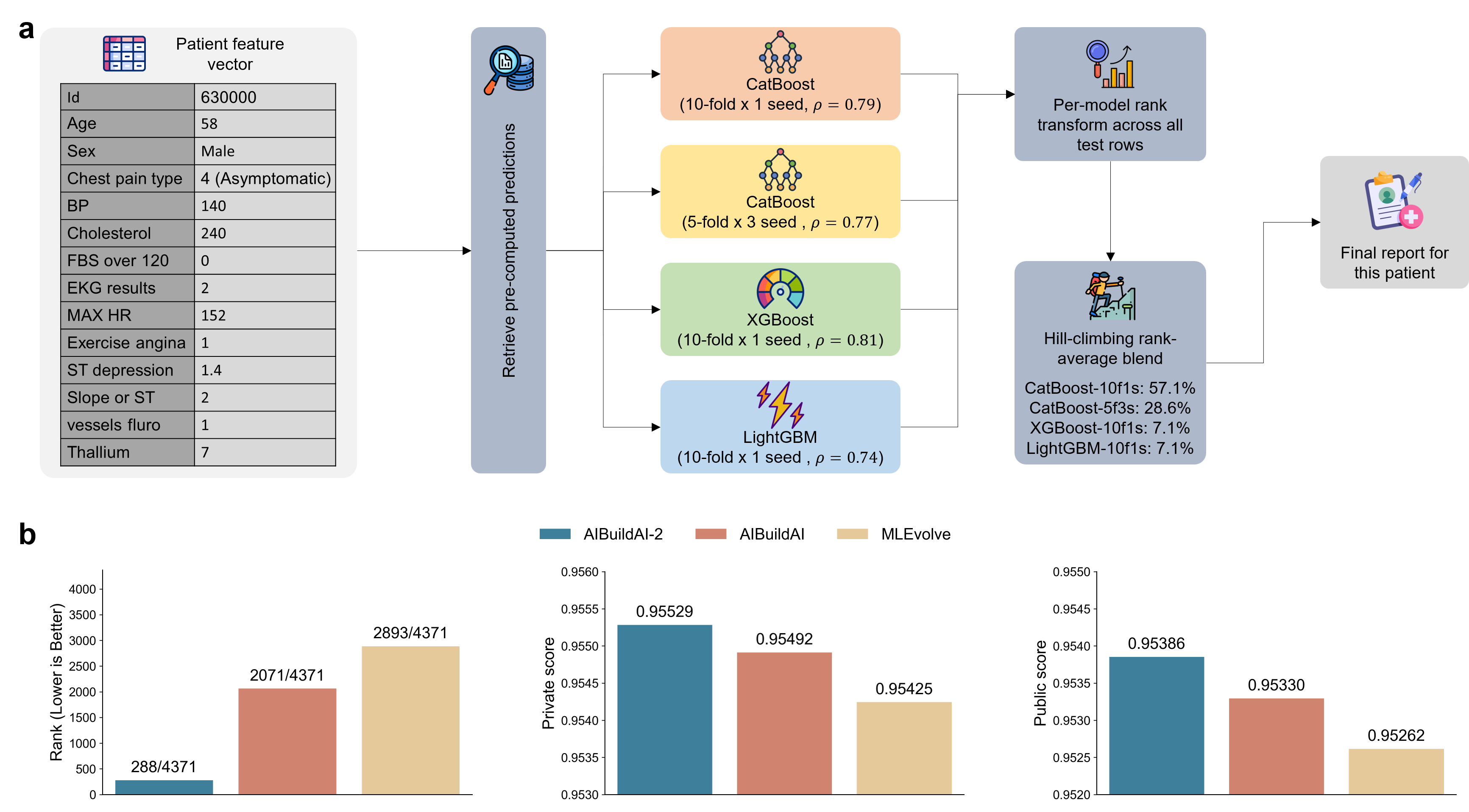}
\caption{\textbf{AIBuildAI-2 ranks among top-tier human experts in a Kaggle heart disease prediction competition.}
\textbf{a,} Final solution in detail. The competition (Kaggle Playground Series S6E2) requires predicting the presence of heart disease from 13 tabular clinical features (e.g., age, blood pressure, cholesterol, maximum heart rate, and ST depression) over 630{,}000 training and 270{,}000 test rows. AIBuildAI-2 produces a heterogeneous gradient-boosted decision-tree ensemble built from four base models, each trained on the raw clinical features with cross-validation: a CatBoost~\cite{prokhorenkova2018catboost} trained with 10-fold cross-validation, a second CatBoost trained with 5-fold cross-validation, an XGBoost~\cite{chen2016XGBoost} trained with 10-fold cross-validation, and a LightGBM~\cite{Ke2017LightGBM} trained with 10-fold cross-validation. Each base model's predictions are rank-transformed across all test rows and combined by a hill-climbing rank-average blend that assigns weights of 57.1\%, 28.6\%, 7.1\%, and 7.1\% to the four models, producing the final per-patient prediction.
\textbf{b,} Competition results. Submissions are ranked by AUC against 4{,}370 human-expert teams. AIBuildAI-2 places 288th out of 4{,}371 (top 6.6\%) with a private AUC of 0.95529 and a public AUC of 0.95386, clearly outperforming two baseline AI development agents, AIBuildAI (private AUC 0.95492, public AUC 0.95330, rank 2{,}071), which relies on web-search-based knowledge, and MLEvolve (private AUC 0.95425, public AUC 0.95262, rank 2{,}893).}
\label{fig:heart-disease}
\end{figure*}

Beyond static benchmarks, we further evaluate AIBuildAI-2 in a live, real-world Kaggle competition against thousands of human-expert teams. The Heart Disease Prediction competition (Kaggle Playground Series S6E2)~\cite{kaggle2026heartdisease}, derived from the canonical UCI Cleveland heart-disease dataset~\cite{detrano1989heartdisease}, requires building a binary classifier that predicts the presence of heart disease from 13 tabular clinical features, including age, blood pressure, cholesterol, maximum heart rate, and ST depression, with 630{,}000 training rows and 270{,}000 test rows, evaluated by AUC on a held-out test set. The competition attracted 4{,}370 teams composed of human AI practitioners and domain experts, providing a stringent test of whether an autonomous AI development agent can match expert human performance on a clinically meaningful prediction task.
\newline
\newline
Fig.~\ref{fig:heart-disease}a depicts the final solution automatically developed by AIBuildAI-2 in detail. AIBuildAI-2 produces a heterogeneous gradient-boosted decision-tree ensemble over four base models that operate directly on the 13 raw clinical features: a CatBoost~\cite{prokhorenkova2018catboost} trained with 10-fold cross-validation, a second CatBoost trained with 5-fold cross-validation, an XGBoost~\cite{chen2016XGBoost} trained with 10-fold cross-validation, and a LightGBM~\cite{Ke2017LightGBM} trained with 10-fold cross-validation. Each base model's predictions are rank-transformed across all test rows on a per-model basis, so that the four models are placed on a common scale insensitive to absolute calibration. The four rank-transformed prediction vectors are combined by a hill-climbing rank-average blend~\cite{caruana2004ensemble} that searches for the convex weight configuration most predictive on out-of-fold data, yielding final weights of 57.1\% on the first CatBoost, 28.6\% on the second CatBoost, 7.1\% on the XGBoost, and 7.1\% on the LightGBM. The blended rank score for each test patient is the per-patient prediction reported on the leaderboard.
\newline
\newline
Without any task-specific customization, this AI model produced by AIBuildAI-2 achieves a private-leaderboard AUC of 0.95529 (public AUC 0.95386), placing 288th out of 4{,}371 teams (the 4{,}370 human-expert teams plus AIBuildAI-2's submission) and ranking in the top 6.6\% of the competition (Fig.~\ref{fig:heart-disease}b). We compare against two baseline AI development agents on the same competition: AIBuildAI~\cite{zhang2026aibuildai} attains a private AUC of 0.95492 (public AUC 0.95330) at rank 2{,}071, and MLEvolve~\cite{Du2025AutoMLGenNF} attains a private AUC of 0.95425 (public AUC 0.95262) at rank 2{,}893. AIBuildAI-2 clearly outperforms both baselines, whose knowledge components are limited to web-search retrieval (AIBuildAI) and a static knowledge base (MLEvolve), and reaches a level competitive with top-tier human experts. This demonstrates the effectiveness of our comprehensive and continually evolving knowledge system in driving real-world AI model development performance.

\subsection*{AIBuildAI-2 generalizes to a blind ADMET drug discovery challenge}

\begin{figure*}[!htb]
\centering
\includegraphics[width=\linewidth]{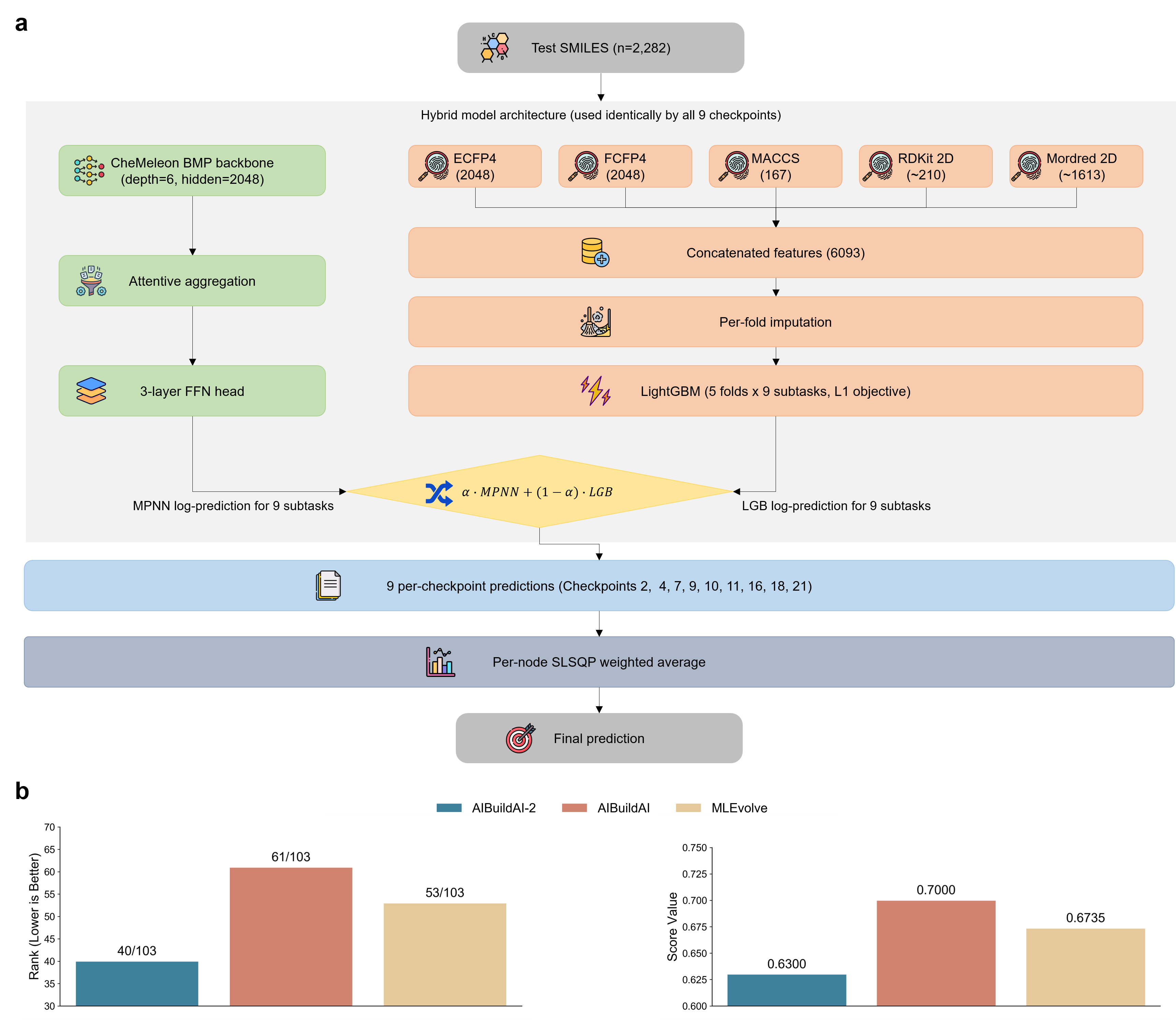}
\caption{\textbf{AIBuildAI-2 outperforms baseline agents in a blind ADMET drug discovery challenge.}
\textbf{a,} Final solution in detail. The OpenADMET ExpansionRx Blind Challenge requires jointly predicting nine ADMET subtasks (e.g., LogD, kinetic solubility, human and mouse liver microsomal intrinsic clearance, Caco-2 apparent permeability, and plasma protein binding) for held-out small molecules from their SMILES strings. AIBuildAI-2 develops a hybrid model that jointly predicts all nine subtasks: a graph branch passes the SMILES string through a CheMeleon~\cite{burns2025chemeleon} bond-message-passing backbone, an attentive aggregation, and a 3-layer feed-forward head, while a tabular branch concatenates ECFP4 and FCFP4 circular fingerprints~\cite{rogers2010ecfp}, MACCS keys, and RDKit~\cite{landrum2024rdkit} and Mordred~\cite{moriwaki2018mordred} 2D descriptors into a single feature vector, applies per-fold mean imputation, and trains a 5-fold LightGBM~\cite{Ke2017LightGBM}. The two branches' predictions are linearly blended with a per-subtask weight to produce predictions for all nine subtasks. The final solution is a per-subtask weighted average over nine independently trained checkpoints of this architecture, with the weights fitted using sequential least squares quadratic programming (SLSQP)~\cite{kraft1988slsqp}, producing the final per-molecule prediction across all nine subtasks.
\textbf{b,} Competition results. Submissions are ranked by mean absolute relative absolute error (MA-RAE) across subtasks. AIBuildAI-2 attains an MA-RAE of 0.6300, ranking 40th out of 103 teams, and substantially outperforms two baseline AI development agents, AIBuildAI (MA-RAE 0.7000, rank 61/103) and MLEvolve (MA-RAE 0.6735, rank 53/103).}
\label{fig:openadmet}
\end{figure*}

We further evaluate AIBuildAI-2 on a domain that differs substantially from MLE-Bench and the heart disease competition: blind prediction of ADMET properties in early-stage drug discovery. ADMET refers to absorption, distribution, metabolism, excretion, and toxicity, the five classes of pharmacokinetic and safety properties that determine whether a candidate small molecule can become a viable drug, and accurately predicting them from chemical structure alone is a long-standing bottleneck in drug discovery~\cite{huang2021tdc}. The OpenADMET ExpansionRx Blind Challenge~\cite{openadmet2026expansionrx} is a community competition in which participants are given a fixed training set and must submit predictions for held-out test molecules whose labels are revealed only after the submission deadline (hence ``blind''). It requires predicting nine pharmacologically relevant ADMET subtasks, including LogD (lipophilicity, octanol-water distribution at physiological pH), kinetic solubility (the concentration of dissolved compound under non-equilibrium aqueous conditions), human and mouse liver microsomal intrinsic clearance (an in-vitro proxy for metabolic stability in each species), Caco-2 apparent permeability (a cell-based proxy for intestinal absorption), and plasma protein binding (the fraction of compound bound to blood proteins, which limits free drug concentration), with each molecule given only as a SMILES string~\cite{weininger1988smiles}. The model is trained on 5{,}326 molecules, in which most subtask labels are sparsely observed and missing not at random, and is evaluated on 2{,}282 held-out molecules. Submissions are scored by mean absolute relative absolute error (MA-RAE) averaged across the nine subtasks on a hidden test set, and 103 teams composed of computational chemists and machine learning researchers participated in the challenge. Compared with the previous two settings, this benchmark stresses the agent's ability to handle a specialized scientific modality (molecular structures), to integrate cheminformatics-specific representations and pretrained models, and to cope with sparse, heterogeneous regression targets.
\newline
\newline
Fig.~\ref{fig:openadmet}a depicts the final solution automatically developed by AIBuildAI-2 in detail. The solution is a hybrid model that jointly predicts all nine ADMET subtasks. A graph branch processes the SMILES string through a CheMeleon~\cite{burns2025chemeleon} bond-message-passing backbone~\cite{gilmer2017mpnn,yang2019chemprop} with a depth of six message-passing steps and a hidden size of 2{,}048, followed by an attentive aggregation~\cite{xiong2020attentivefp} and a 3-layer feed-forward head, producing a log-space MPNN prediction for each subtask. In parallel, a tabular branch concatenates ECFP4 and FCFP4 circular fingerprints~\cite{rogers2010ecfp}, MACCS keys~\cite{durant2002maccs}, and 2D descriptors from RDKit~\cite{landrum2024rdkit} and Mordred~\cite{moriwaki2018mordred} into a 6{,}093-dimensional feature vector, applies per-fold mean imputation to handle missing values, and trains a 5-fold LightGBM~\cite{Ke2017LightGBM} model. The two branches are combined by a linear blend whose weight is tuned per subtask on out-of-fold data, yielding a single checkpoint. The final solution is a per-subtask weighted average over nine independently trained checkpoints of this architecture, where the per-subtask weights are constrained to be non-negative and to sum to one and are fitted on out-of-fold predictions using sequential least squares quadratic programming (SLSQP)~\cite{kraft1988slsqp}.
\newline
\newline
Without any task-specific customization, this automatically developed model achieves an MA-RAE of 0.6300, ranking 40th out of 103 teams (top 38.8\%) on the leaderboard (Fig.~\ref{fig:openadmet}b). We compare against two baseline AI development agents on the same competition: AIBuildAI~\cite{zhang2026aibuildai} attains an MA-RAE of 0.7000 (rank 61/103), and MLEvolve~\cite{Du2025AutoMLGenNF} attains an MA-RAE of 0.6735 (rank 53/103). AIBuildAI-2 substantially outperforms both baselines, whose knowledge components are limited to web-search retrieval (AIBuildAI) and a static knowledge base (MLEvolve), and reaches a level competitive with human ADMET experts. This demonstrates the effectiveness of our comprehensive and continually evolving knowledge system on a domain that requires specialized molecular knowledge, exposing the agent to drug-discovery-specific resources such as cheminformatics fingerprints and graph neural network architectures for molecules that are insufficiently covered by generic LLM parametric knowledge or unconstrained web search.

\section*{Discussion}

Across three evaluation settings spanning general machine-learning benchmarks, a real-world clinical prediction task, and a scientifically demanding drug discovery problem, AIBuildAI-2 attains a level of performance comparable to expert human practitioners: it ranks first on MLE-Bench, places in the top 6.6\% of 4{,}370 human-expert teams on the heart disease prediction competition, and ranks 40th out of 103 teams on the OpenADMET ExpansionRx Blind Challenge, all without task-specific customization. Taken together, these results demonstrate that AIBuildAI-2 provides a reliable, generally applicable route for researchers, including those without deep AI engineering expertise, to automatically build competitive AI models for both general benchmark tasks and domain-specific scientific problems.
\newline
\newline
The strong performance of AIBuildAI-2 can be attributed in large part to its comprehensive, hierarchical, and continually evolving knowledge system, which grounds each design and implementation decision in concrete external expertise rather than the parametric memory of the underlying LLM alone. Building high-performing AI models requires substantial practical engineering know-how, such as which architectures suit a given data modality, which preprocessing stabilizes training, and which failure modes to anticipate during debugging, much of which is fragmented across competition write-ups, technical blogs, library documentation, and open-source code, and is rarely covered in depth by generalist LLMs. By organizing around 1{,}000 such documents under a human-authored L1 index of around 30 topical categories, with each L1 value providing a high-level knowledge instruction together with an L2 index of pointers to the underlying low-level knowledge documents, AIBuildAI-2 makes this expertise systematically accessible to its sub-agents, while dynamic context loading brings only the relevant slice of knowledge into the context window at each step. After each run, an L2 builder turns the completed task into a structured takeaway appended to the L2 lookup under the most relevant category, and an L1 builder refines the corresponding L1 value, allowing the agent to accumulate experience and progressively improve.
\newline
\newline
AIBuildAI-2 also differs fundamentally from prior LLM-based autonomous AI development agents in how it leverages external knowledge. Many existing agents~\cite{jiang2025aide,yang2025rdagent} rely entirely on the parametric knowledge of their underlying LLMs, which is static, often outdated, and sparse on engineering know-how. Others augment this with on-the-fly web search~\cite{toledo2025airesearchagentsmachine,zhang2026aibuildai}, but search results are noisy, inconsistent across queries, and offer no mechanism to consolidate what has been learned over time. A more recent line of work~\cite{Du2025AutoMLGenNF} couples the agent with a fixed corpus of pre-collected documents, providing more reliable context but only over a narrow, static slice of AI development knowledge. Concurrent work~\cite{zhu2026mlmaster2} also explores letting an AI development agent accumulate knowledge from its own past runs, but its self-evolution draws solely on the agent's own experience; AIBuildAI-2 instead jointly evolves the knowledge system from both the agent's experience and newly emerging web content, and uses a substantively different hierarchical design. AIBuildAI-2 addresses these limitations simultaneously through breadth (an L2 lookup of around 1{,}000 curated low-level knowledge documents), hierarchical accessibility (a human-authored L1 index whose values combine high-level knowledge instructions with pointers into L2, retrieved on demand), and self-evolution (post-run takeaways processed by L1 and L2 builders that update both layers), which together drive its consistent gains across MLE-Bench, the heart disease prediction competition, and the OpenADMET drug discovery challenge.
\newline
\newline
A limitation of AIBuildAI-2 is that, while its knowledge system evolves from both completed runs and newly published web content, the surrounding harness, that is, the prompts and code that define each sub-agent's behavior and their orchestration, remains fixed. Recent agentic systems have begun to explore evolving the harness itself~\cite{hu2025ads,rosser2025agentbreeder}, suggesting a natural extension in which AIBuildAI-2 learns from its own runs not only what knowledge to retain but also how to prompt and structure its sub-agents. However, harness-level evolution is intrinsically more difficult than knowledge-level evolution: small changes to prompts or control flow can interact unpredictably with the LLM and with the search dynamics across repositories, leading to high variance and instability across runs. Stabilizing such updates while preserving the gains demonstrated here is a substantive open problem and an important direction for future research.
\newline
\newline
Beyond raw task performance, the design of AIBuildAI-2 carries a broader implication. Because its knowledge system is curated, structured, and continually evolving, solving AI tasks yields, as a by-product, a human-readable corpus of AI development knowledge that is automatically collected, validated through use, and refined over time. This corpus makes the model-development process interpretable: by inspecting the L1 instructions and L2 documents, researchers can trace which strategies the agent considered, which succeeded, and why, turning an otherwise opaque automated process into an auditable source of AI development insight. The implication is most pronounced in AI-for-science~\cite{wang2023scientific}, where general AI development strategies must be adapted to domain-specific data, constraints, and evaluation protocols, and where such adaptations are sparsely documented because they demand deep expertise in both AI and the relevant scientific discipline. As AIBuildAI-2 operates, its knowledge system accumulates precisely these adaptations, offering a route to systematically distill, share, and build upon AI-for-science expertise that is otherwise confined to a small number of specialists.

\section*{References}
\bibliographystyle{zHenriquesLab-StyleBib}
\bibliography{ref}

\begin{thebibliography}{53}
\providecommand{\natexlab}[1]{#1}
\providecommand{\url}[1]{\texttt{#1}}
\expandafter\ifx\csname urlstyle\endcsname\relax
  \providecommand{\doi}[1]{doi: #1}\else
  \providecommand{\doi}{doi: \begingroup \urlstyle{rm}\Url}\fi

\bibitem[He et~al.(2016)He, Zhang, Ren, and Sun]{he2016deep}
Kaiming He, Xiangyu Zhang, Shaoqing Ren, and Jian Sun.
\newblock Deep residual learning for image recognition.
\newblock In Lourdes Agapito, Tamara Berg, Jana Kosecka, and Lihi Zelnik-Manor, editors, \emph{Proceedings of the IEEE Conference on Computer Vision and Pattern Recognition}, pages 770--778, 2016.

\bibitem[Otter et~al.(2020)Otter, Medina, and Kalita]{otter2020survey}
Daniel~W Otter, Julian~R Medina, and Jugal~K Kalita.
\newblock A survey of the usages of deep learning for natural language processing.
\newblock \emph{IEEE transactions on neural networks and learning systems}, 32\penalty0 (2):\penalty0 604--624, 2020.

\bibitem[Singhal et~al.(2023)Singhal, Azizi, Tu, Mahdavi, Wei, Chung, Scales, Tanwani, Cole-Lewis, Pfohl, Payne, Seneviratne, Gamble, Kelly, Babiker, Sch{\"a}rli, Chowdhery, Mansfield, Demner-Fushman, Ag{\"u}era~y Arcas, Webster, Corrado, Matias, Chou, Gottweis, Tomasev, Liu, Rajkomar, Barral, Semturs, Karthikesalingam, and Natarajan]{singhal2023clinical}
Karan Singhal, Shekoofeh Azizi, Tao Tu, S.~Sara Mahdavi, Jason Wei, Hyung~Won Chung, Nathan Scales, Ajay Tanwani, Heather Cole-Lewis, Stephen Pfohl, Perry Payne, Martin Seneviratne, Paul Gamble, Chris Kelly, Abubakr Babiker, Nathanael Sch{\"a}rli, Aakanksha Chowdhery, Philip Mansfield, Dina Demner-Fushman, Blaise Ag{\"u}era~y Arcas, Dale Webster, Greg~S. Corrado, Yossi Matias, Katherine Chou, Juraj Gottweis, Nenad Tomasev, Yun Liu, Alvin Rajkomar, Joelle Barral, Christopher Semturs, Alan Karthikesalingam, and Vivek Natarajan.
\newblock Large language models encode clinical knowledge.
\newblock \emph{Nature}, 620\penalty0 (7972):\penalty0 172--180, 2023.
\newblock \doi{10.1038/s41586-023-06291-2}.

\bibitem[Singhal et~al.(2025)Singhal, Tu, Gottweis, Sayres, Wulczyn, Amin, Hou, Clark, Pfohl, Cole-Lewis, Neal, Rashid, Schaekermann, Wang, Dash, Chen, Shah, Lachgar, Mansfield, Prakash, Green, Dominowska, Ag{\"u}era~y Arcas, Toma{\v{s}}ev, Liu, Wong, Semturs, Mahdavi, Barral, Webster, Corrado, Matias, Azizi, Karthikesalingam, and Natarajan]{singhal2025medqa}
Karan Singhal, Tao Tu, Juraj Gottweis, Rory Sayres, Ellery Wulczyn, Mohamed Amin, Le~Hou, Kevin Clark, Stephen~R. Pfohl, Heather Cole-Lewis, Darlene Neal, Qazi~Mamunur Rashid, Mike Schaekermann, Amy Wang, Dev Dash, Jonathan~H. Chen, Nigam~H. Shah, Sami Lachgar, Philip~Andrew Mansfield, Sushant Prakash, Bradley Green, Ewa Dominowska, Blaise Ag{\"u}era~y Arcas, Nenad Toma{\v{s}}ev, Yun Liu, Renee Wong, Christopher Semturs, S.~Sara Mahdavi, Joelle~K. Barral, Dale~R. Webster, Greg~S. Corrado, Yossi Matias, Shekoofeh Azizi, Alan Karthikesalingam, and Vivek Natarajan.
\newblock Toward expert-level medical question answering with large language models.
\newblock \emph{Nature Medicine}, 31\penalty0 (3):\penalty0 943--950, 2025.
\newblock \doi{10.1038/s41591-024-03423-7}.

\bibitem[Jumper et~al.(2021)Jumper, Evans, Pritzel, Green, Figurnov, Ronneberger, Tunyasuvunakool, Bates, {\v{Z}}{\'i}dek, Potapenko, Bridgland, Meyer, Kohl, Ballard, Cowie, Romera-Paredes, Nikolov, Jain, Adler, Back, Petersen, Reiman, Clancy, Zielinski, Steinegger, Pacholska, Berghammer, Bodenstein, Silver, Vinyals, Senior, Kavukcuoglu, Kohli, and Hassabis]{jumper2021alphafold}
John Jumper, Richard Evans, Alexander Pritzel, Tim Green, Michael Figurnov, Olaf Ronneberger, Kathryn Tunyasuvunakool, Russ Bates, Augustin {\v{Z}}{\'i}dek, Anna Potapenko, Alex Bridgland, Clemens Meyer, Simon A.~A. Kohl, Andrew~J. Ballard, Andrew Cowie, Bernardino Romera-Paredes, Stanislav Nikolov, Rishub Jain, Jonas Adler, Trevor Back, Stig Petersen, David Reiman, Ellen Clancy, Michal Zielinski, Martin Steinegger, Michalina Pacholska, Tamas Berghammer, Sebastian Bodenstein, David Silver, Oriol Vinyals, Andrew~W. Senior, Koray Kavukcuoglu, Pushmeet Kohli, and Demis Hassabis.
\newblock Highly accurate protein structure prediction with {AlphaFold}.
\newblock \emph{Nature}, 596\penalty0 (7873):\penalty0 583--589, 2021.
\newblock \doi{10.1038/s41586-021-03819-2}.

\bibitem[Wang et~al.(2023)Wang, Fu, Du, Gao, Huang, Liu, Chandak, Liu, Van~Katwyk, Deac, Anandkumar, Bergen, Gomes, Ho, Kohli, Lasenby, Leskovec, Liu, Manrai, Marks, Ramsundar, Song, Sun, Tang, Veli{\v{c}}kovi{\'c}, Welling, Zhang, Coley, Bengio, and Zitnik]{wang2023scientific}
Hanchen Wang, Tianfan Fu, Yuanqi Du, Wenhao Gao, Kexin Huang, Ziming Liu, Payal Chandak, Shengchao Liu, Peter Van~Katwyk, Andreea Deac, Anima Anandkumar, Karianne Bergen, Carla~P. Gomes, Shirley Ho, Pushmeet Kohli, Joan Lasenby, Jure Leskovec, Tie-Yan Liu, Arjun Manrai, Debora Marks, Bharath Ramsundar, Le~Song, Jimeng Sun, Jian Tang, Petar Veli{\v{c}}kovi{\'c}, Max Welling, Linfeng Zhang, Connor~W. Coley, Yoshua Bengio, and Marinka Zitnik.
\newblock Scientific discovery in the age of artificial intelligence.
\newblock \emph{Nature}, 620\penalty0 (7972):\penalty0 47--60, 2023.
\newblock \doi{10.1038/s41586-023-06221-2}.

\bibitem[Stokes et~al.(2020)Stokes, Yang, Swanson, Jin, Cubillos-Ruiz, Donghia, MacNair, French, Carfrae, Bloom-Ackermann, Tran, Chiappino-Pepe, Badran, Andrews, Chory, Church, Brown, Jaakkola, Barzilay, and Collins]{stokes2020antibiotic}
Jonathan~M. Stokes, Kevin Yang, Kyle Swanson, Wengong Jin, Andres Cubillos-Ruiz, Nina~M. Donghia, Craig~R. MacNair, Shawn French, Lindsey~A. Carfrae, Zohar Bloom-Ackermann, Victoria~M. Tran, Anush Chiappino-Pepe, Ahmed~H. Badran, Ian~W. Andrews, Emma~J. Chory, George~M. Church, Eric~D. Brown, Tommi~S. Jaakkola, Regina Barzilay, and James~J. Collins.
\newblock A deep learning approach to antibiotic discovery.
\newblock \emph{Cell}, 180\penalty0 (4):\penalty0 688--702.e13, 2020.
\newblock \doi{10.1016/j.cell.2020.01.021}.

\bibitem[Wong et~al.(2023)Wong, Zheng, Valeri, Donghia, Anahtar, Omori, Li, Cubillos-Ruiz, Krishnan, Jin, Manson, Friedrichs, Helbig, Hajian, Fiejtek, Wagner, Soutter, Earl, Stokes, Renner, and Collins]{wong2023antibiotics}
Felix Wong, Erica~J. Zheng, Jacqueline~A. Valeri, Nina~M. Donghia, Melis~N. Anahtar, Satotaka Omori, Alicia Li, Andres Cubillos-Ruiz, Aarti Krishnan, Wengong Jin, Abigail~L. Manson, Jens Friedrichs, Ralf Helbig, Behnoush Hajian, Dawid~K. Fiejtek, Florence~F. Wagner, Holly~H. Soutter, Ashlee~M. Earl, Jonathan~M. Stokes, Lars~D. Renner, and James~J. Collins.
\newblock Discovery of a structural class of antibiotics with explainable deep learning.
\newblock \emph{Nature}, 626\penalty0 (7997):\penalty0 177--185, 2023.
\newblock \doi{10.1038/s41586-023-06887-8}.

\bibitem[Turing(2007)]{turing2007computing}
Alan~M Turing.
\newblock \emph{Computing machinery and intelligence}, pages 23--65.
\newblock Springer, 2007.

\bibitem[Jordan and Mitchell(2015)]{jordan2015machine}
Michael~I Jordan and Tom~M Mitchell.
\newblock Machine learning: Trends, perspectives, and prospects.
\newblock \emph{Science}, 349\penalty0 (6245):\penalty0 255--260, 2015.

\bibitem[Ghahramani(2015)]{ghahramani2015probabilistic}
Zoubin Ghahramani.
\newblock Probabilistic machine learning and artificial intelligence.
\newblock \emph{Nature}, 521\penalty0 (7553):\penalty0 452--459, 2015.

\bibitem[Biamonte et~al.(2017)Biamonte, Wittek, Pancotti, Rebentrost, Wiebe, and Lloyd]{biamonte2017quantum}
Jacob Biamonte, Peter Wittek, Nicola Pancotti, Patrick Rebentrost, Nathan Wiebe, and Seth Lloyd.
\newblock Quantum machine learning.
\newblock \emph{Nature}, 549\penalty0 (7671):\penalty0 195--202, 2017.

\bibitem[Bergstra and Bengio(2012)]{bergstra2012random}
James Bergstra and Yoshua Bengio.
\newblock Random search for hyper-parameter optimization.
\newblock \emph{Journal of machine learning research}, 13\penalty0 (2), 2012.

\bibitem[Sculley et~al.(2015)Sculley, Holt, Golovin, Davydov, Phillips, Ebner, Chaudhary, Young, Crespo, and Dennison]{sculley2015hidden}
D~Sculley, Gary Holt, Daniel Golovin, Eugene Davydov, Todd Phillips, Dietmar Ebner, Vinay Chaudhary, Michael Young, Jean-Francois Crespo, and Dan Dennison.
\newblock Hidden technical debt in machine learning systems.
\newblock In Corinna Cortes, Neil~D. Lawrence, Daniel~D. Lee, Masashi Sugiyama, and Roman Garnett, editors, \emph{Proceedings of the International Conference on Neural Information Processing Systems}, volume~2, pages 2503--2511, 2015.

\bibitem[Hutter et~al.(2019)Hutter, Kotthoff, and Vanschoren]{hutter2019automated}
Frank Hutter, Lars Kotthoff, and Joaquin Vanschoren.
\newblock \emph{Automated machine learning: methods, systems, challenges}.
\newblock Springer, 2019.

\bibitem[Aldoseri et~al.(2023)Aldoseri, Al-Khalifa, and Hamouda]{aldoseri2023re}
Abdulaziz Aldoseri, Khalifa~N Al-Khalifa, and Abdel~Magid Hamouda.
\newblock Re-thinking data strategy and integration for artificial intelligence: concepts, opportunities, and challenges.
\newblock \emph{Applied Sciences}, 13\penalty0 (12):\penalty0 7082, 2023.

\bibitem[Yang et~al.(2024)Yang, Zhang, Gichoya, Katabi, and Ghassemi]{yang2024limits}
Yuzhe Yang, Haoran Zhang, Judy~W Gichoya, Dina Katabi, and Marzyeh Ghassemi.
\newblock The limits of fair medical imaging ai in real-world generalization.
\newblock \emph{Nature medicine}, 30\penalty0 (10):\penalty0 2838--2848, 2024.

\bibitem[Toledo et~al.(2025)Toledo, Hambardzumyan, Josifoski, HAZRA, Baldwin, Audran-Reiss, Kuchnik, Magka, Jiang, Lupidi, Lupu, Raileanu, Shavrina, Niu, Gagnon-Audet, Shvartsman, Sodhani, Miller, Charnalia, Dunfield, Wu, Stenetorp, Cancedda, Foerster, and Bachrach]{toledo2025airesearchagentsmachine}
Edan Toledo, Karen Hambardzumyan, Martin Josifoski, RISHI HAZRA, Nicolas Baldwin, Alexis Audran-Reiss, Michael Kuchnik, Despoina Magka, Minqi Jiang, Alisia~Maria Lupidi, Andrei Lupu, Roberta Raileanu, Tatiana Shavrina, Kelvin Niu, Jean-Christophe Gagnon-Audet, Michael Shvartsman, Shagun Sodhani, Alexander~H Miller, Abhishek Charnalia, Derek Dunfield, Carole-Jean Wu, Pontus Stenetorp, Nicola Cancedda, Jakob~Nicolaus Foerster, and Yoram Bachrach.
\newblock {AI} research agents for machine learning: Search, exploration, and generalization in {MLE}-bench.
\newblock 2025.

\bibitem[Du et~al.(2025)Du, Yan, Jiang, Yuan, Hu, Li, He, Zhang, and Bai]{Du2025AutoMLGenNF}
Shangheng Du, Xiangchao Yan, Dengyang Jiang, Jiakang Yuan, Yusong Hu, Xin Li, Liang He, Bo~Zhang, and Lei Bai.
\newblock {AutoMLGen}: Navigating fine-grained optimization for coding agents.
\newblock \emph{arXiv preprint arXiv:2510.08511}, 2025.

\bibitem[Jiang et~al.(2025)Jiang, Schmidt, Srikanth, Xu, Kaplan, Jacenko, and Wu]{jiang2025aide}
Zhengyao Jiang, Dominik Schmidt, Dhruv Srikanth, Dixing Xu, Ian Kaplan, Deniss Jacenko, and Yuxiang Wu.
\newblock {AIDE}: {AI-Driven} exploration in the space of code.
\newblock \emph{arXiv preprint arXiv:2502.13138}, 2025.

\bibitem[Yang et~al.(2025)Yang, Yang, Fang, Zhang, Wang, Xian, Li, Li, Xu, Li, Pan, Zhang, Liu, Shen, Chen, and Bian]{yang2025rdagent}
Xu~Yang, Xiao Yang, Shikai Fang, Yifei Zhang, Jian Wang, Bowen Xian, Qizheng Li, Jingyuan Li, Minrui Xu, Yuante Li, Haoran Pan, Yuge Zhang, Weiqing Liu, Yelong Shen, Weizhu Chen, and Jiang Bian.
\newblock {R\&D-Agent}: An {LLM-Agent} framework towards autonomous data science.
\newblock \emph{arXiv preprint arXiv:2505.14738}, 2025.

\bibitem[Zhang et~al.(2026)Zhang, Qin, Cao, Zhang, and Xie]{zhang2026aibuildai}
Ruiyi Zhang, Peijia Qin, Qi~Cao, Li~Zhang, and Pengtao Xie.
\newblock {AIBuildAI}: An {AI} agent for automatically building {AI} models.
\newblock \emph{arXiv preprint arXiv:2604.14455}, 2026.

\bibitem[Chan et~al.(2025)Chan, Chowdhury, Jaffe, Aung, Sherburn, Mays, Starace, Liu, Maksin, Patwardhan, Madry, and Weng]{chan2025mlebench}
Jun~Shern Chan, Neil Chowdhury, Oliver Jaffe, James Aung, Dane Sherburn, Evan Mays, Giulio Starace, Kevin Liu, Leon Maksin, Tejal Patwardhan, Aleksander Madry, and Lilian Weng.
\newblock {MLE-bench}: Evaluating machine learning agents on machine learning engineering, 2025.
\newblock International Conference on Learning Representations (ICLR).

\bibitem[{Kaggle}()]{kaggle}
{Kaggle}.
\newblock {Kaggle}: Your machine learning and data science community.
\newblock \url{https://www.kaggle.com}.
\newblock Accessed: 2026-05-20.

\bibitem[OpenAI(2026)]{mlebench_commit_2026}
OpenAI.
\newblock Mle-bench leaderboard (commit c5631ba).
\newblock \url{https://github.com/openai/mle-bench/tree/c5631ba61ceeb0573235a6ce209db435327a1e84}, 2026.
\newblock Accessed: 2026-03-18.

\bibitem[Lewis et~al.(2020)Lewis, Perez, Piktus, Petroni, Karpukhin, Goyal, K{\"u}ttler, Lewis, Yih, Rockt{\"a}schel, Riedel, and Kiela]{lewis2020rag}
Patrick Lewis, Ethan Perez, Aleksandra Piktus, Fabio Petroni, Vladimir Karpukhin, Naman Goyal, Heinrich K{\"u}ttler, Mike Lewis, Wen-tau Yih, Tim Rockt{\"a}schel, Sebastian Riedel, and Douwe Kiela.
\newblock Retrieval-augmented generation for knowledge-intensive {NLP} tasks.
\newblock \emph{Advances in Neural Information Processing Systems}, 2020.

\bibitem[Singh et~al.(2025)Singh, Ehtesham, Kumar, Khoei, and Vasilakos]{singh2025agenticrag}
Aditi Singh, Abul Ehtesham, Saket Kumar, Tala~Talaei Khoei, and Athanasios~V. Vasilakos.
\newblock Agentic retrieval-augmented generation: A survey on agentic {RAG}.
\newblock \emph{arXiv preprint arXiv:2501.09136}, 2025.

\bibitem[Chen et~al.(2026)Chen, Dalvi~Mishra, Nam, Meng, Pfister, and Yoon]{chen2026mars}
Jiefeng Chen, Bhavana Dalvi~Mishra, Jaehyun Nam, Rui Meng, Tomas Pfister, and Jinsung Yoon.
\newblock {MARS}: Modular agent with reflective search for automated {AI} research.
\newblock \emph{arXiv preprint arXiv:2602.02660}, 2026.

\bibitem[Li et~al.(2025)Li, Wu, Ge, Chong, Hou, Cao, Ju, Wu, Li, Zhang, Feng, Zhao, Qiu, Yang, Zhang, Zhu, Sun, Sun, Yan, Liu, Yin, and Shen]{li2025fmagent}
Annan Li, Chufan Wu, Zengle Ge, Yee~Hin Chong, Zhinan Hou, Lizhe Cao, Cheng Ju, Jianmin Wu, Huaiming Li, Haobo Zhang, Shenghao Feng, Mo~Zhao, Fengzhi Qiu, Rui Yang, Mengmeng Zhang, Wenyi Zhu, Yingying Sun, Quan Sun, Shunhao Yan, Danyu Liu, Dawei Yin, and Dou Shen.
\newblock The {FM} agent.
\newblock \emph{arXiv preprint arXiv:2510.26144}, 2025.

\bibitem[Liu et~al.(2025)Liu, Cai, Zhu, Zheng, Chen, Wen, Wang, E, and Chen]{liu2025mlmaster}
Zexi Liu, Yuzhu Cai, Xinyu Zhu, Yujie Zheng, Runkun Chen, Ying Wen, Yanfeng Wang, Weinan E, and Siheng Chen.
\newblock {ML-Master}: Towards {AI-for-AI} via integration of exploration and reasoning.
\newblock \emph{arXiv preprint arXiv:2506.16499}, 2025.

\bibitem[Nadafian et~al.(2026)Nadafian, Mohammadshahi, and Yazdani]{nadaf2026kapso}
Alireza Nadafian, Alireza Mohammadshahi, and Majid Yazdani.
\newblock {KAPSO}: A knowledge-grounded framework for autonomous program synthesis and optimization.
\newblock \emph{arXiv preprint arXiv:2601.21526}, 2026.

\bibitem[Team et~al.(2025)Team, Zhang, Feng, Yan, Yuan, Ma, Hu, Yu, He, Huang, et~al.]{team2025internagent}
InternAgent Team, Bo~Zhang, Shiyang Feng, Xiangchao Yan, Jiakang Yuan, Runmin Ma, Yusong Hu, Zhiyin Yu, Xiaohan He, Songtao Huang, et~al.
\newblock {InternAgent}: When agent becomes the scientist---building closed-loop system from hypothesis to verification.
\newblock \emph{arXiv preprint arXiv:2505.16938}, 2025.

\bibitem[Prokhorenkova et~al.(2018)Prokhorenkova, Gusev, Vorobev, Dorogush, and Gulin]{prokhorenkova2018catboost}
Liudmila Prokhorenkova, Gleb Gusev, Aleksandr Vorobev, Anna~Veronika Dorogush, and Andrey Gulin.
\newblock {CatBoost}: Unbiased boosting with categorical features.
\newblock \emph{Advances in Neural Information Processing Systems}, 31, 2018.

\bibitem[Chen and Guestrin(2016)]{chen2016XGBoost}
Tianqi Chen and Carlos Guestrin.
\newblock Xgboost: A scalable tree boosting system, 2016.

\bibitem[Ke et~al.(2017)Ke, Meng, Finley, Wang, Chen, Ma, Ye, and Liu]{Ke2017LightGBM}
Guolin Ke, Qi~Meng, Thomas Finley, Taifeng Wang, Wei Chen, Weidong Ma, Qiwei Ye, and Tie-Yan Liu.
\newblock Lightgbm: a highly efficient gradient boosting decision tree, 2017.

\bibitem[{Kaggle}(2026)]{kaggle2026heartdisease}
{Kaggle}.
\newblock Predicting heart disease ({Playground} {Series} {S6E2}).
\newblock \url{https://www.kaggle.com/competitions/playground-series-s6e2/overview}, 2026.
\newblock Accessed: 2026-04-29.

\bibitem[Detrano et~al.(1989)Detrano, Janosi, Steinbrunn, Pfisterer, Schmid, Sandhu, Guppy, Lee, and Froelicher]{detrano1989heartdisease}
Robert Detrano, Andras Janosi, Walter Steinbrunn, Matthias Pfisterer, Johann-Jakob Schmid, Sarbjit Sandhu, Kern~H. Guppy, Stella Lee, and Victor Froelicher.
\newblock International application of a new probability algorithm for the diagnosis of coronary artery disease.
\newblock \emph{The American Journal of Cardiology}, 64\penalty0 (5):\penalty0 304--310, 1989.
\newblock \doi{10.1016/0002-9149(89)90524-9}.

\bibitem[Caruana et~al.(2004)Caruana, Niculescu-Mizil, Crew, and Ksikes]{caruana2004ensemble}
Rich Caruana, Alexandru Niculescu-Mizil, Geoff Crew, and Alex Ksikes.
\newblock Ensemble selection from libraries of models.
\newblock \emph{Proceedings of the Twenty-first International Conference on Machine Learning}, page~18, 2004.
\newblock \doi{10.1145/1015330.1015432}.

\bibitem[Burns et~al.(2025)Burns, Zalte, Abreu, Sieg, Feldmann, Mathea, and Green]{burns2025chemeleon}
Jackson~W. Burns, Akshat~Shirish Zalte, Charlles R.~A. Abreu, Jochen Sieg, Christian Feldmann, Miriam Mathea, and William~H. Green.
\newblock Deep learning foundation models from classical molecular descriptors.
\newblock \emph{arXiv preprint arXiv:2506.15792}, 2025.

\bibitem[Rogers and Hahn(2010)]{rogers2010ecfp}
David Rogers and Mathew Hahn.
\newblock Extended-connectivity fingerprints.
\newblock \emph{Journal of Chemical Information and Modeling}, 50\penalty0 (5):\penalty0 742--754, 2010.
\newblock \doi{10.1021/ci100050t}.

\bibitem[Landrum and {The RDKit Contributors}(2024)]{landrum2024rdkit}
Greg Landrum and {The RDKit Contributors}.
\newblock {RDKit}: Open-source cheminformatics software.
\newblock \url{https://www.rdkit.org}, 2024.
\newblock Accessed: 2026-04-29.

\bibitem[Moriwaki et~al.(2018)Moriwaki, Tian, Kawashita, and Takagi]{moriwaki2018mordred}
Hirotomo Moriwaki, Yu-Shi Tian, Norihito Kawashita, and Tatsuya Takagi.
\newblock Mordred: A molecular descriptor calculator.
\newblock \emph{Journal of Cheminformatics}, 10\penalty0 (1):\penalty0 4, 2018.
\newblock \doi{10.1186/s13321-018-0258-y}.

\bibitem[Kraft(1988)]{kraft1988slsqp}
Dieter Kraft.
\newblock A software package for sequential quadratic programming.
\newblock Tech. Rep. DFVLR-FB 88-28, {DFVLR}, Institut f\"ur Dynamik der Flugsysteme, Oberpfaffenhofen, Germany, 1988.

\bibitem[Huang et~al.(2021)Huang, Fu, Gao, Zhao, Roohani, Leskovec, Coley, Xiao, Sun, and Zitnik]{huang2021tdc}
Kexin Huang, Tianfan Fu, Wenhao Gao, Yue Zhao, Yusuf Roohani, Jure Leskovec, Connor~W. Coley, Cao Xiao, Jimeng Sun, and Marinka Zitnik.
\newblock Therapeutics data commons: Machine learning datasets and tasks for drug discovery and development.
\newblock \emph{Advances in Neural Information Processing Systems Datasets and Benchmarks Track}, 2021.

\bibitem[{OpenADMET}(2026)]{openadmet2026expansionrx}
{OpenADMET}.
\newblock {OpenADMET} {ExpansionRx} blind challenge.
\newblock \url{https://huggingface.co/spaces/openadmet/OpenADMET-ExpansionRx-Challenge}, 2026.
\newblock Accessed: 2026-04-29.

\bibitem[Weininger(1988)]{weininger1988smiles}
David Weininger.
\newblock {SMILES}, a chemical language and information system. 1. introduction to methodology and encoding rules.
\newblock \emph{Journal of Chemical Information and Computer Sciences}, 28\penalty0 (1):\penalty0 31--36, 1988.
\newblock \doi{10.1021/ci00057a005}.

\bibitem[Gilmer et~al.(2017)Gilmer, Schoenholz, Riley, Vinyals, and Dahl]{gilmer2017mpnn}
Justin Gilmer, Samuel~S. Schoenholz, Patrick~F. Riley, Oriol Vinyals, and George~E. Dahl.
\newblock Neural message passing for quantum chemistry.
\newblock \emph{Proceedings of the 34th International Conference on Machine Learning (ICML)}, pages 1263--1272, 2017.

\bibitem[Yang et~al.(2019)Yang, Swanson, Jin, Coley, Eiden, Gao, Guzman-Perez, Hopper, Kelley, Mathea, Palmer, Settels, Jaakkola, Jensen, and Barzilay]{yang2019chemprop}
Kevin Yang, Kyle Swanson, Wengong Jin, Connor Coley, Philipp Eiden, Hua Gao, Angel Guzman-Perez, Timothy Hopper, Brian Kelley, Miriam Mathea, Andrew Palmer, Volker Settels, Tommi Jaakkola, Klavs Jensen, and Regina Barzilay.
\newblock Analyzing learned molecular representations for property prediction.
\newblock \emph{Journal of Chemical Information and Modeling}, 59\penalty0 (8):\penalty0 3370--3388, 2019.
\newblock \doi{10.1021/acs.jcim.9b00237}.

\bibitem[Xiong et~al.(2020)Xiong, Wang, Liu, Zhong, Wan, Li, Li, Luo, Chen, Jiang, and Zheng]{xiong2020attentivefp}
Zhaoping Xiong, Dingyan Wang, Xiaohong Liu, Feisheng Zhong, Xiaozhe Wan, Xutong Li, Zhaojun Li, Xiaomin Luo, Kaixian Chen, Hualiang Jiang, and Mingyue Zheng.
\newblock Pushing the boundaries of molecular representation for drug discovery with the graph attention mechanism.
\newblock \emph{Journal of Medicinal Chemistry}, 63\penalty0 (16):\penalty0 8749--8760, 2020.
\newblock \doi{10.1021/acs.jmedchem.9b00959}.

\bibitem[Durant et~al.(2002)Durant, Leland, Henry, and Nourse]{durant2002maccs}
Joseph~L. Durant, Burton~A. Leland, Douglas~R. Henry, and James~G. Nourse.
\newblock Reoptimization of {MDL} keys for use in drug discovery.
\newblock \emph{Journal of Chemical Information and Computer Sciences}, 42\penalty0 (6):\penalty0 1273--1280, 2002.
\newblock \doi{10.1021/ci010132r}.

\bibitem[Zhu et~al.(2026)Zhu, Cai, Liu, Zheng, Wang, Ye, Zhang, Zhang, E, Chen, and Wang]{zhu2026mlmaster2}
Xinyu Zhu, Yuzhu Cai, Zexi Liu, Bingyang Zheng, Cheng Wang, Rui Ye, Yuzhi Zhang, Linfeng Zhang, Weinan E, Siheng Chen, and Yanfeng Wang.
\newblock Toward ultra-long-horizon agentic science: Cognitive accumulation for machine learning engineering.
\newblock \emph{arXiv preprint arXiv:2601.10402}, 2026.

\bibitem[Hu et~al.(2025)Hu, Lu, and Clune]{hu2025ads}
Shengran Hu, Cong Lu, and Jeff Clune.
\newblock Automated design of agentic systems.
\newblock \emph{International Conference on Learning Representations}, 2025.

\bibitem[Rosser and Foerster(2025)]{rosser2025agentbreeder}
J.~Rosser and Jakob~Nicolaus Foerster.
\newblock Agentbreeder: Mitigating the ai safety risks of multi-agent scaffolds via self-improvement.
\newblock \emph{Advances in Neural Information Processing Systems}, 2025.

\end{thebibliography}


\begin{thebibliography}{13}
\providecommand{\natexlab}[1]{#1}
\providecommand{\url}[1]{\texttt{#1}}
\expandafter\ifx\csname urlstyle\endcsname\relax
  \providecommand{\doi}[1]{doi: #1}\else
  \providecommand{\doi}{doi: \begingroup \urlstyle{rm}\Url}\fi

\bibitem[Mart\'{i} et~al.(2013)Mart\'{i}, Resende, and Ribeiro]{marti2013multistart}
Rafael Mart\'{i}, Mauricio G.~C. Resende, and Celso~C. Ribeiro.
\newblock Multi-start methods for combinatorial optimization.
\newblock \emph{European Journal of Operational Research}, 226\penalty0 (1):\penalty0 1--8, 2013.

\bibitem[Yao et~al.(2023)Yao, Zhao, Yu, Du, Shafran, Narasimhan, and Cao]{yao2023react}
Shunyu Yao, Jeffrey Zhao, Dian Yu, Nan Du, Izhak Shafran, Karthik Narasimhan, and Yuan Cao.
\newblock {ReAct}: Synergizing reasoning and acting in language models.
\newblock \emph{International Conference on Learning Representations}, 2023.

\bibitem[{Agent Skills}(2025)]{agentskills2025}
{Agent Skills}.
\newblock Agent skills: An open standard for extending {AI} agent capabilities.
\newblock \url{https://agentskills.io/home}, 2025.
\newblock Accessed: 2026-05-20.

\bibitem[Anthropic(2025)]{anthropic2025skills}
Anthropic.
\newblock Introducing {Agent} {Skills}.
\newblock \url{https://claude.com/blog/skills}, 2025.
\newblock Accessed: 2026-04-27.

\bibitem[{OpenAI}(2025)]{openai2025skills}
{OpenAI}.
\newblock {OpenAI} skills.
\newblock \url{https://openai.com/academy/skills/}, 2025.
\newblock Accessed: 2026-05-20.

\bibitem[Paszke et~al.(2019)Paszke, Gross, Massa, Lerer, Bradbury, Chanan, Killeen, Lin, Gimelshein, Antiga, Desmaison, K{\"o}pf, Yang, DeVito, Raison, Tejani, Chilamkurthy, Steiner, Fang, Bai, and Chintala]{paszke2019pytorch}
Adam Paszke, Sam Gross, Francisco Massa, Adam Lerer, James Bradbury, Gregory Chanan, Trevor Killeen, Zeming Lin, Natalia Gimelshein, Luca Antiga, Alban Desmaison, Andreas K{\"o}pf, Edward Yang, Zach DeVito, Martin Raison, Alykhan Tejani, Sasank Chilamkurthy, Benoit Steiner, Lu~Fang, Junjie Bai, and Soumith Chintala.
\newblock {PyTorch}: An imperative style, high-performance deep learning library.
\newblock \emph{Advances in Neural Information Processing Systems}, 2019.

\bibitem[{Hugging Face}()]{huggingface}
{Hugging Face}.
\newblock {Hugging Face}: The ai community building the future.
\newblock \url{https://huggingface.co}.
\newblock Accessed: 2026-05-20.

\bibitem[Pedregosa et~al.(2011)Pedregosa, Varoquaux, Gramfort, Michel, Thirion, Grisel, Blondel, Prettenhofer, Weiss, Dubourg, Vanderplas, Passos, Cournapeau, Brucher, Perrot, and Duchesnay]{pedregosa2011scikit}
Fabian Pedregosa, Ga{\"e}l Varoquaux, Alexandre Gramfort, Vincent Michel, Bertrand Thirion, Olivier Grisel, Mathieu Blondel, Peter Prettenhofer, Ron Weiss, Vincent Dubourg, Jake Vanderplas, Alexandre Passos, David Cournapeau, Matthieu Brucher, Matthieu Perrot, and {\'E}douard Duchesnay.
\newblock Scikit-learn: Machine learning in {P}ython.
\newblock \emph{Journal of Machine Learning Research}, 12:\penalty0 2825--2830, 2011.

\bibitem[Wolf et~al.(2020)Wolf, Debut, Sanh, Chaumond, Delangue, Moi, Cistac, Rault, Louf, Funtowicz, Davison, Shleifer, von Platen, Ma, Jernite, Plu, Xu, Le~Scao, Gugger, Drame, Lhoest, and Rush]{wolf2020transformers}
Thomas Wolf, Lysandre Debut, Victor Sanh, Julien Chaumond, Clement Delangue, Anthony Moi, Pierric Cistac, Tim Rault, R{\'e}mi Louf, Morgan Funtowicz, Joe Davison, Sam Shleifer, Patrick von Platen, Clara Ma, Yacine Jernite, Julien Plu, Canwen Xu, Teven Le~Scao, Sylvain Gugger, Mariama Drame, Quentin Lhoest, and Alexander~M. Rush.
\newblock Transformers: State-of-the-art natural language processing.
\newblock \emph{Proceedings of the 2020 Conference on Empirical Methods in Natural Language Processing: System Demonstrations}, pages 38--45, 2020.
\newblock \doi{10.18653/v1/2020.emnlp-demos.6}.

\bibitem[{GitHub}()]{github}
{GitHub}.
\newblock {GitHub}: Build software better, together.
\newblock \url{https://github.com}.
\newblock Accessed: 2026-05-20.

\bibitem[{arXiv}()]{arxiv}
{arXiv}.
\newblock {arXiv}.org: Open-access archive for scholarly articles.
\newblock \url{https://arxiv.org}.
\newblock Accessed: 2026-05-20.

\bibitem[Broder(1997)]{broder1997minhash}
Andrei~Z. Broder.
\newblock On the resemblance and containment of documents.
\newblock \emph{Proceedings of the Compression and Complexity of Sequences}, pages 21--29, 1997.
\newblock \doi{10.1109/SEQUEN.1997.666900}.

\bibitem[Anthropic(2026)]{anthropic2026claude47}
Anthropic.
\newblock Claude opus 4.7 system card.
\newblock \url{https://anthropic.com/claude-opus-4-7-system-card}, 2026.
\newblock Accessed: 2026-04-27.

\end{thebibliography}

\section*{Methods}

\subsection*{Problem formulation}

We formalize automated AI model development as the task of constructing a runnable AI solution from a task description and a dataset. The input consists of (i) a natural language description of the AI task and its evaluation protocol, and (ii) a dataset partitioned into training and test splits. The required output is an inference artifact, namely trained model checkpoints together with an inference script that produces predictions on unseen test data. The objective is to maximize the task's evaluation metric (e.g., AUC, accuracy, MA-RAE) on the held-out test split.
\newline
\newline
AIBuildAI-2 approaches this task with a multi-start local search strategy~\citeMethod{marti2013multistart}: rather than committing to a single modeling approach, AIBuildAI-2 maintains $n$ solution repositories $\{\text{repo}_i\}_{i=1}^n$ in parallel, each pursuing a fundamentally different modeling strategy and isolated within its own workspace, and iteratively refines them. Each repository $\text{repo}_i$ contains all artifacts needed to produce the final inference output, organized under a common schema $(\text{plan}_i, \text{code}_i, \text{config}_i, \text{result}_i)$: a textual plan $\text{plan}_i$ describing the modeling strategy, code $\text{code}_i$ implementing the training and inference pipelines, configuration $\text{config}_i$ specifying hyperparameters and other training settings, and execution results $\text{result}_i$ from training and validation runs. Throughout the search, these repositories are continually updated as their artifacts are revised; the surviving repositories are finally aggregated, by selection or ensembling, into a single inference script with its associated checkpoints that constitutes the required output of the task defined above.
\newline
\newline
Concretely, AIBuildAI-2 is the LLM-based agent that carries out this parallel update of repositories and produces the final aggregation. Following the standard agent paradigm~\citeMethod{yao2023react}, an LLM is invoked iteratively in a control loop where each call may either return a final response or request a tool invocation, whose execution result is appended to the context for subsequent calls; a specific agent is then fully specified by its prompt and its tool set~\cite{zhang2026aibuildai}. The tool set of AIBuildAI-2 comprises the repository-scoped tools $\text{read}_i$, $\text{write}_i$, and $\text{execute}_i$ for inspecting artifacts, modifying files, and running code in each workspace, together with a knowledge-retrieval tool $\text{query}$ described later, and these tools allow the agent to conduct experiments and update repository artifacts in place. At each step, AIBuildAI-2 selects a repository and an action that refines its artifacts, observes the resulting $\text{result}_i$, and updates its strategy accordingly; improving repositories are pursued further and underperforming ones are halted until the budget is exhausted, and the strongest candidates are then selected or ensembled into the final inference output.

\subsection*{Hierarchical agent architecture}

In principle, a single LLM-based agent equipped with the full tool set above could carry out the entire workflow described in the previous section. In practice, however, having one agent jointly manage parallel search, modeling decisions, code implementation, and training-time tuning across all repositories quickly overflows its context window, which both inflates token cost and degrades reasoning quality. We therefore decompose the complete workflow into sub-tasks handled by different sub-agents, structuring AIBuildAI-2 internally as a hierarchy of four LLM-based sub-agents, each with a focused prompt and a restricted tool set: a top-level manager that orchestrates the search across repositories, and three specialized workers (designer, coder, tuner) that, on a given repository, respectively handle modeling strategy, code implementation, and training-time tuning. The manager's tool set consists of the repository-scoped sub-agent tools ($\text{designer}_i$, $\text{coder}_i$, $\text{tuner}_i$), the cross-repository $\text{read}_i$ tools, and $\text{query}$; tool invocations are asynchronous, so the manager can issue parallel calls and overlap design, implementation, and training across repositories. For each repository $i$, the designer sub-agent produces or revises $\text{plan}_i$: when $\text{plan}_i$ is absent it generates an initial modeling, data, and training strategy informed by knowledge retrieval and external search; when $\text{plan}_i$ and $\text{result}_i$ both exist it diagnoses failure modes such as underfitting, overfitting, or optimization instability and updates $\text{plan}_i$ accordingly. The coder sub-agent translates $\text{plan}_i$ into executable code, producing an initial $\text{code}_i$ together with $\text{config}_i$, performing a short verification run, and iteratively debugging until the pipeline executes end-to-end. The tuner sub-agent runs training and tunes performance by analyzing logs, metrics, and checkpoints to detect slow convergence, overfitting, or instability, then updating $\text{config}_i$ and launching new runs; short preliminary runs are used to discard ineffective configurations early, and promising ones are extended to full training within the budget. Surrounding this core loop, a setup agent prepares the shared software environment at initialization, and an aggregator agent selects or ensembles the strongest repositories at termination to produce the final inference output.

\subsection*{Knowledge system}\label{sec:kb}

The knowledge system underpinning AIBuildAI-2 has two levels: a top level $L_1$ of around 30 high-level knowledge instructions, one per topical category of AI model development, and a bottom level $L_2$ of around 1{,}000 low-level knowledge documents that each L1 instruction points to for more specific knowledge tailored to narrower scenarios within the category. The L1 index is a compact document, authored by human AI experts, that defines the taxonomy of the knowledge system: each entry consists of a category key $k$ together with a short description, of a few lines, summarizing the type of AI development knowledge that falls under that category. The categories fall into two groups: modality- or task-specific knowledge (e.g., vision, NLP, time series, tabular, molecular) and AI-modeling-strategy-specific knowledge tied to a particular technique (e.g., feature engineering, ensembling). For each category key $k$, the L1 value $L_1[k]$ has two components: a high-level knowledge instruction and a per-category L2 index. The high-level knowledge instruction is a self-contained set of guidelines on how to approach category $k$, written in the spirit of agent skills~\citeMethod{agentskills2025}, such as Anthropic Skills~\citeMethod{anthropic2025skills} and OpenAI Skills~\citeMethod{openai2025skills}, so that it can be loaded directly into an LLM-based agent's context. The per-category L2 index lists, for each available document under $k$, the document identifier $j$ paired with a short description summarizing what the document contains, allowing the agent to decide whether to retrieve a given document without first loading its full text. The actual low-level knowledge documents pointed to by these entries are stored in the L2 lookup, keyed by both category and document: $L_2[k][j]$ holds a single clean, self-contained document distilled from a group of related sources that all address one specific, narrowly scoped task or method within category $k$, such as competition write-ups, technical blog posts, open-source code, and library documentation, or distilled from the working directory of a previous AIBuildAI-2 run.
\newline
\newline
The main agent and its sub-agents access this knowledge system through a single $\text{query}$ tool that implements dynamic context loading. The L1 index is included in every agent invocation by default, so the agent always has a global view of available guidance. The tool exposes two call signatures. A category-level call $\text{query}(k) = L_1[k]$ loads the high-level knowledge instruction for category $k$ together with its L2 index of pointers, giving the agent both the condensed guidance for $k$ and the option to drill down further. After this category-level call returns, the agent can either proceed using the high-level knowledge instruction alone, when the condensed guidance in $L_1[k]$ is sufficient for the current decision, or, when it judges that more specific knowledge tailored to a narrower scenario is needed (for example, a feature extraction recipe for a particular type of data or tricks for improving a specific type of model), consult the per-category L2 index inside $L_1[k]$ to identify a relevant document by its short description and issue a document-level call $\text{query}(k, j) = L_2[k][j]$ to load the corresponding low-level knowledge document. This loading strategy ensures that only the knowledge actually needed for each decision is brought into the context, substantially reducing context consumption.

\subsection*{Knowledge builder agents}

The two value layers are constructed and maintained by two LLM-based agents that operate offline before deployment and, during continual evolution, also after each completed run. The L2 builder accepts two types of input: (1) a group of similar pieces of evidence about AI model development, such as multiple competition write-ups, blog posts, or open-source code repositories that cover the same task or method, and (2) a complete AIBuildAI-2 working directory from a finished run. Given an input together with the L1 index, the agent is prompted to (i) classify the input into the most relevant category $k$, since no category label is given a priori; (ii) convert it into a clean, self-contained document, by consolidating and de-duplicating multiple noisy web sources into a single normalized document or by summarizing a working directory into a takeaway aimed at improving future runs on similar tasks; and (iii) produce a short description for retrieval. The result is the low-level knowledge document $L_2[k][j]$, stored under $k$ with its description. These tasks demand strong reasoning over heterogeneous artifacts, motivating an LLM-based agent rather than a fixed pipeline.
\newline
\newline
The L1 builder takes (i) zero or one existing $L_1[k]$ for a target category $k$ and (ii) one or more low-level knowledge documents under $k$, and emits an updated $L_1[k]$ comprising a refreshed high-level knowledge instruction and an updated per-category L2 index. The high-level knowledge instruction is prompted to be clean, self-contained, and to clearly instruct an agent on how to solve the AI tasks in category $k$ or to carry out the specific AI method that $k$ represents. When no L1 value exists yet for $k$, the L1 builder writes $L_1[k]$ from scratch using the low-level knowledge documents most relevant to $k$, identified by clustering document embeddings and verified via an LLM relevance check; when $L_1[k]$ already exists, the same agent edits the existing instruction in place and extends its L2 index. The human-expert-curated L1 index itself is held fixed across both the initial build and the subsequent evolution process.

\subsection*{Knowledge system initialization}

To bootstrap $L_2$, we crawl AI-development-related sources from Kaggle~\cite{kaggle} competition forums, deep learning framework blogs (PyTorch~\citeMethod{paszke2019pytorch}, Hugging Face~\citeMethod{huggingface}), official documentation of widely used AI libraries (scikit-learn~\citeMethod{pedregosa2011scikit}, Transformers~\citeMethod{wolf2020transformers}), open-source code repositories on GitHub~\citeMethod{github}, and selected arXiv~\citeMethod{arxiv} collections. We then apply a combination of MinHash-based deduplication~\citeMethod{broder1997minhash} and an LLM-based relevance classifier to filter out low-quality or off-topic crawled content and to group together sources that cover the same task or method. The L2 builder is run on each such group of similar sources to produce a single structured low-level knowledge document $L_2[k][j]$ stored under the most relevant category $k$, together with its description. This procedure yields around 1{,}000 L2 documents distributed across around 30 categories. The L1 builder is then run on each category $k$ in the human-authored L1 index, taking as input the low-level knowledge documents under $k$ and emitting $L_1[k]$ from scratch. The result is the initial knowledge system that AIBuildAI-2 starts from at deployment time, with each L1 value providing a high-level knowledge instruction over its category and a per-category L2 index pointing to the underlying low-level knowledge documents.

\subsection*{Knowledge system evolution}\label{sec:kb_evolve}

The knowledge system continues to evolve after deployment through two complementary update streams. The first stream folds AIBuildAI-2's own experience back into the knowledge system: after each completed task, the L2 builder is run on the AIBuildAI-2 working directory. The second stream periodically incorporates newly published web content: as new competition write-ups, blog posts, library updates, and code repositories appear, the L2 builder is run on each new group of similar sources, so $L_2$ continues to track the state of the field rather than being frozen at the bootstrap snapshot. Each invocation in either stream produces an additional low-level knowledge document $L_2[k^\star][j^\star]$, stored under the most relevant category $k^\star$ and appended to $L_2[k^\star]$ alongside the bootstrap corpus.
\newline
\newline
In both update streams, once a new low-level knowledge document $L_2[k^\star][j^\star]$ is added under category $k^\star$, the L1 builder is re-invoked in evolution mode on $k^\star$ with the existing $L_1[k^\star]$ and the new document as input, and emits an updated $L_1[k^\star]$ in which the high-level knowledge instruction is refined to incorporate new strategies or annotate known pitfalls and the per-category L2 index is extended to point to the new document, while the original structure and length budget of $L_1[k^\star]$ are preserved. This update procedure keeps the knowledge system well-formed and lightweight while ensuring that it accumulates increasingly comprehensive and battle-tested guidance derived from both the agent's own experience and the broader AI development community.

\subsection*{Experimental settings}

For the MLE-Bench experiments, evaluation follows the MLE-Bench protocol~\cite{chan2025mlebench} and baseline numbers are taken from the official MLE-Bench leaderboard~\cite{mlebench_commit_2026}. AIBuildAI-2 uses Claude Opus 4.7~\citeMethod{anthropic2026claude47} as the backbone LLM for all agents, with a sampling temperature of 1.0, initializes $n=7$ parallel solution repositories per task, and runs with a budget of 24 hours on a Linux x86-64 machine with 24 vCPUs, 256\,GB of RAM, and one NVIDIA A100 GPU.
For the heart disease prediction and OpenADMET ExpansionRx Blind Challenge experiments, all three methods (AIBuildAI-2, AIBuildAI, and MLEvolve~\cite{Du2025AutoMLGenNF}) are evaluated under matched settings: Claude Opus 4.7~\citeMethod{anthropic2026claude47} as the backbone LLM, a sampling temperature of 1.0, and a budget of 24 hours on a Linux x86-64 machine with 24 vCPUs, 256\,GB of RAM, and one NVIDIA A100 GPU. For each of these two tasks, AIBuildAI-2 and AIBuildAI each initialize $n=7$ parallel solution repositories.

\section*{Data availability}
The MLE-Bench benchmark data used in this study are publicly available from the MLE-Bench repository at \url{https://github.com/openai/mle-bench}. The heart disease prediction competition data are publicly available at \url{https://www.kaggle.com/competitions/playground-series-s6e2/overview}, and the OpenADMET ExpansionRx Blind Challenge data are publicly available at \url{https://huggingface.co/spaces/openadmet/OpenADMET-ExpansionRx-Challenge}. The solution generated by AIBuildAI-2 for the heart disease prediction competition is available at \url{https://github.com/aibuildai/AI-Build-AI/tree/main/tasks/playground-series-s6e2}, and the solution generated by AIBuildAI-2 for the OpenADMET ExpansionRx Blind Challenge is available at \url{https://drive.google.com/drive/folders/1IC07fa3_OyDkRwGqNYpSDzjcE0PYaw1N?usp=sharing}.

\section*{Code availability}
The AIBuildAI-2 software is publicly available at \url{https://github.com/aibuildai/AI-Build-AI}, together with detailed instructions for using the system.

\section*{Author contributions}
R.Z., P.Q., Q.C., and L.Z. contributed to conceptualization, methodology, software, investigation, analysis, writing---original draft, and writing---review and editing. P.X. contributed to conceptualization, methodology, investigation, analysis, writing---original draft, and writing---review and editing.

\section*{Competing interests}
The authors declare no competing interests.
\newline

\bibliographystyleMethod{zHenriquesLab-StyleBib}
\bibliographyMethod{ref}

\end{document}